# Airfoil Shape Optimization using Deep Q-Network

**Siddharth Rout**

*Department of Mechanical Engineering*
*Indian Institute of Technology - Madras*
*India*
*bnr.siddharthrout@gmail.com*

**Prof. Chao-An Lin**

*Department of Power Mechanical Engineering*
*National Tsing Hua University*
*Taiwan*
*calin@pme.nthu.edu.tw*

## ABSTRACT

The document explores the feasibility of using reinforcement learning for drag minimization and lift maximization of standard two-dimensional airfoils. Deep Q-network (DQN) is used over Markov's decision process (MDP) to learn the optimal shape by learning the best changes to the initial shape. The airfoil profile is generated by using Bezier control points. The drag and lift values are calculated from coefficient of pressure values along the profile generated using Xfoil potential flow solver. The positions of control points perpendicular to the chord line are changed to obtain the optimal shape. **text**[1]

## 1 Introduction

Standard airfoils are defined by fixing few of the characteristics like maximum thickness, maximum camber and position of maximum thickness while shape of the airfoil surface is not defined. Shape of airfoils can be optimized for low drag or high lift to drag ratio. Typically, averaged drag coefficient and lift coefficient are used as cost function for shape optimization. Formulation of the cost function would be a complex function of many parameters involving operating parameters and points on surface defining airfoil shape. This makes the optimization difficult to converge and stochastic techniques are realized to be better methods for their robustness. Stochastic methods have been quite popular to avoid the problems of inability to achieve pareto optimality in multi-objective design optimization. Methods like genetic algorithm, simulated annealing, swarm algorithm and stochastic hill climbing are often chosen for the basket of choices. For airfoil shape optimization, evolutionary methods are the most popular methods among others. The computation cost involved in all the above techniques is very high due to the complexity of the problem. The document deals with developing a reinforcement learning (RL) based method for different airfoil optimizations.

### 1.1 Reinforcement Learning

Reinforcement learning (RL) is a sub-set of machine learning paradigm where the software learns its job from its own actions. The software actually learns to take the sequence of actions that produces the best result. Typically, the algorithms are maximization algorithms based on Markov decision process (MDP) to maximize the cumulative reward. The RL techniques depends on exploration-exploitation trade-off to find the optimal action sequences over complex high-dimensional domain. The capability of various RL techniques depends on ability to optimize complex problems and ability to store best actions for range of states in complex environments. RL has been widely used for computer programs to learn playing games. Practical applications of RL are evident in the field of controls and robotics.

The airfoil optimization problem has been modelled as a RL problem. The proposed environment state is defined by the set of y-coordinates defining the profile for a set of x-coordinates. This makes the environment a large dimensional continuous space. To accommodate a wide range of environment space and produce the best action a deep neural network, called deep Q-network (DQN) is used to approximate q-values for every possible state. A large number of points are required to define the shape of airfoil. This creates a set of problems which includes increased complexity of optimization, difficulty of approximating q-values and roughness of profile. To tackle these problems Bezier curves are used to generate the profile with much fewer control points.

The action space is defined as a set of decimal values that denote various step sizes to change the position of a control point in the direction perpendicular to the chord line. It contains both positive and negative decimals.

Not all of the control points are changed, some of them are not changed to fix the constraints like maximum thickness and continuity at the joints of two Bezier curves. An episode is defined as a N-step process where N is the number of changeable control points. The episode starts with the first changeable control point changing the first control point by one of the actions from the action space. Consequently, the next control point is selected and an action is taken from the action space. These steps continue till the last action is taken in the last control point.

## 1.2    Profile Generation

The profile is defined by a large number (100-300) points but using them directly is problematic, so Bezier curves are taken to generate same number of profile points using very few (5-20) control points. Bezier curve is a parametrized curve defined by linear combination of control points weighed by Bernstein polynomials of parameter variable ($t$) to approximate highly complicated curve. The number of control points is n + 1 where n is the degree of Bernstein polynomial ($B_{n,i}$). The curve passes through the first and last control point while it does not pass through other control points.

To keep the constraints of an airfoil intact the first and last control points are fixed and, accordingly, the number of curves is fixed for constraints and leading and trailing edges. For NACA 0012 symmetric airfoil, the thickness at 30% of chord line is kept fixed. For ease of interpolation 4 curves are taken with 5 control points ($P_i$) each are taken. The slope of a curve ($C'$) can be obtained by derivative of parameterized curve. To ensure the differentiability at the top and bottom joints of the airfoil the neighboring control points are fixed to be same as that of the maximum thickness. So, the number of changeable control points for optimization are 3 per curve.

$$B_{n,i}(t) = \frac{n!}{i!\,(n-i)!}\, t^i (1-t)^{n-i}$$

$$C(t) = \sum_{i=0}^{n} B_{n,i}(t) P_i$$

$$C'(t) = n \sum_{i=0}^{n-1} B_{n-1,i}(t)(P_{i+1} - P_i)$$

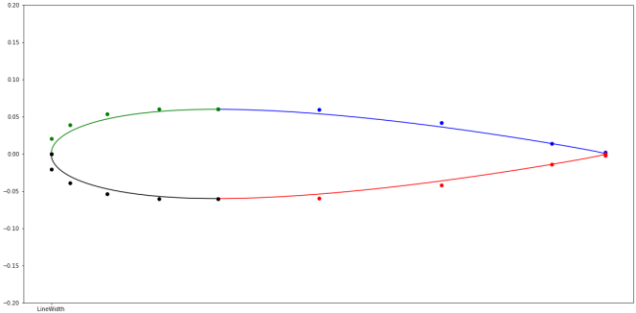

**Fig. 1: Dots represents control points for respective curves**

For asymmetric airfoils a greater number of control points may be required depending on the complexity of approximation.

## 1.3    XFOIL

XFOIL is a FORTRAN based potential flow solver with interactive interface, first developed by Prof. Mark Drela in 1980s at MIT. The latest version (6.99 – December 2013) of the program is upgraded and translated into C++ in collaboration with Harold Youngren. It is widely used for sub-sonic flow simulation around isolated airfoils. Despite being vintage, XFOIL is still widely used for inverse problems and design analysis of two-dimensional airfoils. It takes the airfoil profile, Reynolds number and Mach number as input to calculate the coefficient of pressure around the airfoil for viscid as well as inviscid flows. The lift and drag profile can be determined from the known pressure distribution around the airfoil.

XFOIL has been used to find the pressure distribution around the airfoil. The drag is calculated as component of force due to pressure along the chord line. The lift is calculated as the component of force due to pressure perpendicular to chord line.

Let, pressure = $P$

Pressure at a point with $C_p = C_p P$

Force ($f$) on discrete element (0.5dLx1) = $0.5 C_p P dL$

Lift Force (L) = $f_y = 0.5 C_p P dL \cos\theta$
Lift Measure (LM) = $0.5 C_p\,(\Delta x_i)$
Aim: Maximize LM
Reward for an action step ($R_b$) = $\Sigma f_y$

Drag Force (D) = $f_x = 0.5 C_p P dL \sin\theta$
Drag Measure (DM) = $0.5 C_p(\Delta y_i)$
Aim: Minimize DM
Reward for an action step ($R_a$) = $-\Sigma f_x - \Sigma f_y$

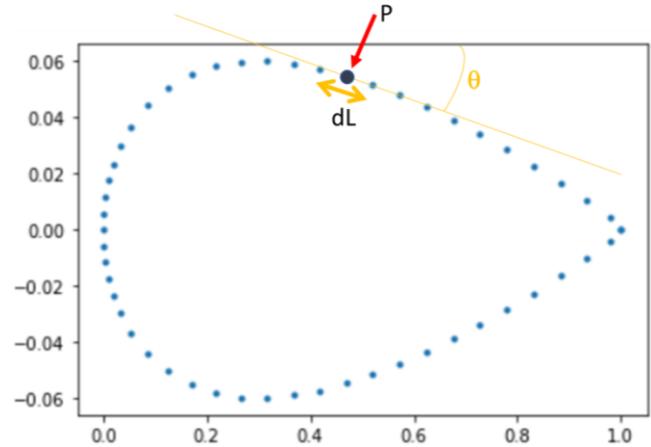

**Fig. 2: Representation for force over a discrete element**

## 2.1    Deep Q-Learning

Q-learning is a model free RL technique. Q stands for quality. It does not require any model specific to the environment space to solve problems. It finds the optimal policy which would maximize the expected cumulative reward over an episode. Using Markov decision process (MDP) it learns the optimal action selection policy given any state of the environment. The policy is usually stored in a table, called Q-table, that maps possible states to possibility (q-values) for various actions from action space. For problems with large environment space, like the environment state defined for airfoil optimization as an array of real values, the number of possible states is extremely large as ideally infinitely many states are possible. To be able to store q-values for such large number of complex states deep neural network is used since neural networks are known for approximating extremely complex functions. These deep networks are called deep Q-network (DQN) since they replace the Q-table.

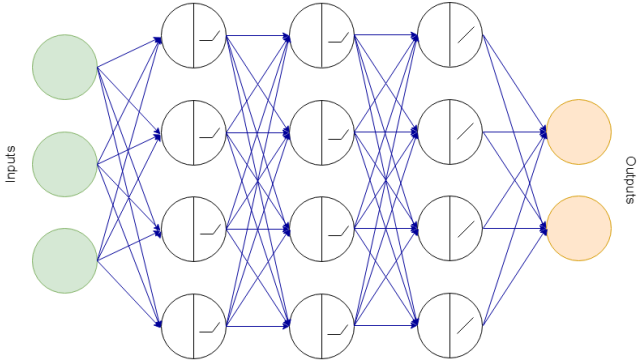

| Q-Table | Action 1 | Action 2 | Action 3 | Action 4 |
|---------|----------|----------|----------|----------|
| State 1 | | | | |
| State 2 | | | | |
| State 3 | | | | |
| State 4 | | | | |
| State 5 | | | | |
| State 6 | | | | |
| State 7 | | | | |
| State 8 | | | | |
| State 9 | | | | |
| State 10 | | | | |

**Fig. 3: Q-Table**

Inputs ... Outputs

**Fig. 4: Deep Q-Network taking environment parameters to define state as input and returns q-values for all possible actions as output**

The DQN is basically a function that calculates the quality of a state action combination. The network is initialized with random weights and trained for arbitrarily chosen q-values for initial state. At each state ($s_t$) in the environment the agent selects an action ($a_t$) to enter new state ($s_{t+1}$) and receives a reward ($r_t$). With each action step the Q-table is can be updated by Bellman Optimality Equation which is used for temporal update in optimal control:

$$Q^*(s_t, a_t) = r(s_t, a_t) + \gamma \max_{a_{t+1}} Q(s_{t+1}, a_{t+1})$$

For DQN the transformed Bellman equation becomes:

$$Q^*(s_t) = r(s_t, a_t) + \gamma \max_{a_{t+1}} Q(s_{t+1})$$

Discount factor ($\gamma$) is the parameter to control the importance of the future rewards over the immediate ones. It ranges from 0 to 1,

depending on importance value. For higher importance to future rewards discount value is kept high.

For Q-Learning the temporal difference of update is used to update Q-tables or DQN. The update equation used is:

$$Q^*(s_t) = Old\ Value + \eta(Target - Old\ Value)$$

Target:      $$r(s_t, a_t) + \gamma \max_{a_{t+1}} Q(s_{t+1})$$

Old Value:      $$Q(s_t)$$

As the DQN learns the temporal difference between target and old value reduces to finally converge. This technique proves to be very efficient in reinforcement learning. $\eta$ is the learning rate for the process. $\eta$(Target – Old Value) is the step size for update.

Since DQN has no training set to learn from, the q-values and action step details are stored for certain number of action steps for training the network batch-wise. This is called experience reply. The actions are generally selected on the basis of balance between exploration and exploitation. Initially the actions are not always selected for maximum q-values in the present state but random actions are taken explore various states to search for global best. As the network learns the algorithm starts selecting the actions corresponding to the maximum q-value more frequently. The exploration-exploitation dilemma is addressed by taking a parameter ($\varepsilon$). For each action step, a random number is generated and if the number is lower than $\varepsilon$ then random action is selected. This is called $\varepsilon$-greedy policy.

## 3    Case

The base case of drag minimization for a symmetric airfoil is considered. The test is conducted over NACA0012 whose constraints are maximum thickness of 12% of chord line and position of maximum thickness occurs at 30% of chord line. The control points are optimized only on upper surface of the airfoil. The lower surface is obtained as reflection of upper surface.

For asymmetric airfoils all the changeable control points are optimized. Even for lift maximization problem all the changeable control points are optimized.

## 4    Results

For drag minimization problem, the profile is divided into four parameterized Bezier curves. The number of control points used for each Bezier curve is five. The coefficient of drag for initial NACA0012 profile produced using XFOIL is found to be 0.00106 for 0° angle of attack. The optimized profile produces a coefficient of drag value of 0.00017 for the same condition using the same solver.

The observations could be noted in the figures below.

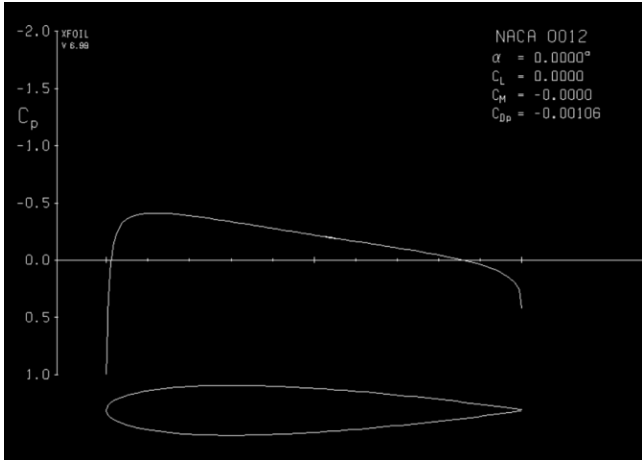

**Fig. 5: Coefficient of pressure for initial NACA 0012 profile generated using XFOIL**

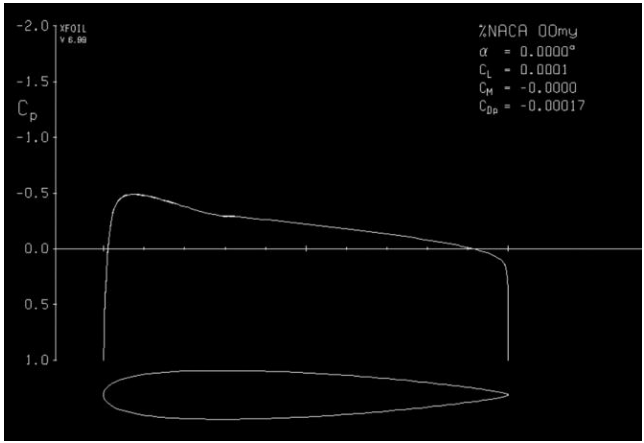

**Fig. 6: Coefficient of pressure for optimized NACA 0012 profile generated using XFOIL**

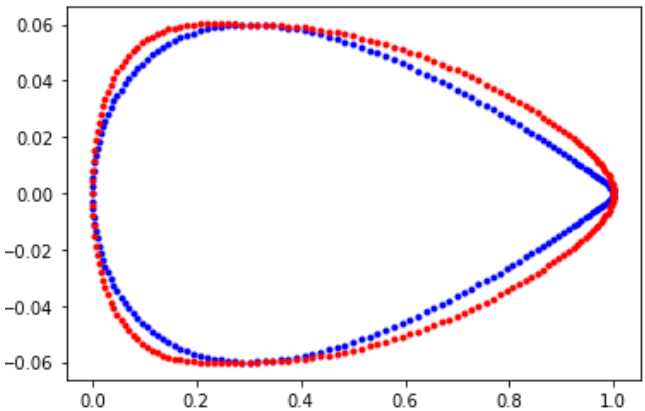

**Fig. 7: Red dots show the drag optimized profile while blue dots show the initial profile for NACA 0012**

For lift maximization at 0° angle of attack, the coefficient of lift calculated using XFOIL is 0.7235.

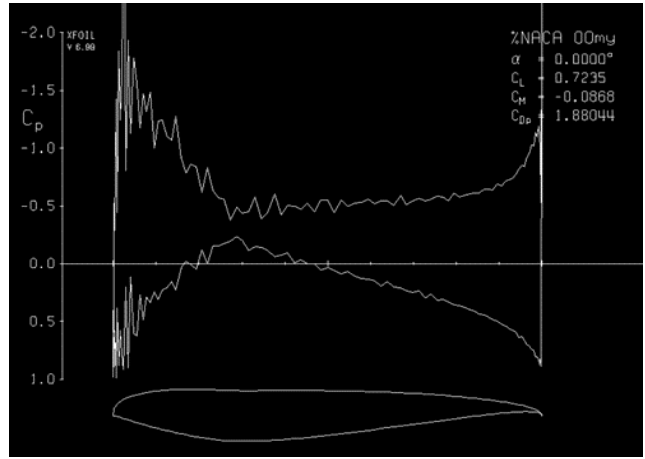

**Fig. 8: Coefficient of pressure for lift maximized NACA 0012 profile generated using XFOIL**

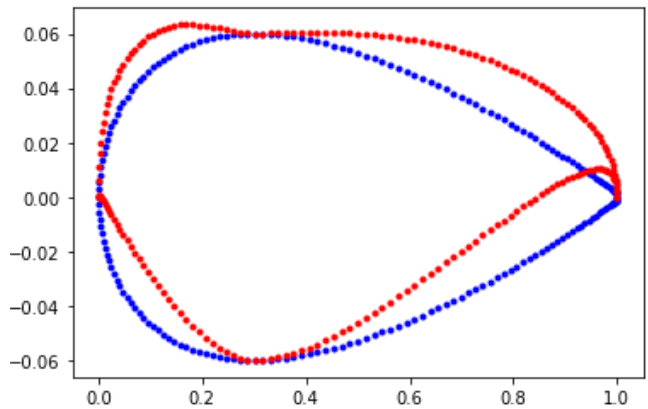

**Fig. 9: Red dots show the lift maximized profile while blue dots show the initial profile for NACA 0012**

## 5    Conclusion

The proposed method works decently for simple cases while optimization for asymmetric airfoils still needs improvement. There are lot of scopes for future developments. The method is not robust as it can not generate the profile from odd shapes.